\documentclass[pdflatex,sn-mathphys-num]{sn-jnl}

\usepackage{graphicx}%
\usepackage{multirow}%
\usepackage{amsmath,amssymb,amsfonts}%
\usepackage{amsthm}%
\usepackage{mathrsfs}%
\usepackage[title]{appendix}%
\usepackage{xcolor}%
\usepackage{textcomp}%
\usepackage{manyfoot}%
\usepackage{booktabs}%
\usepackage{algorithm}%
\usepackage{algorithmicx}%
\usepackage{algpseudocode}%
\usepackage{listings}%

\theoremstyle{thmstyleone}%
\theoremstyle{thmstyletwo}%

\theoremstyle{thmstylethree}%

\begin{document}

\title[Article Title]{A Physics-Informed Neural Network Framework for Elastodynamic Wave Propagation in Bimaterial Systems}


\author[1]{\fnm{Sonal Ankush} \sur{Chibire}}\email{Z2018284@students.niu.edu}

\author[1]{\fnm{Jenn-Terng} \sur{Gau}}\email{jgau@niu.edu}

\author*[1]{\fnm{Bo} \sur{Zhang}}\email{bzhang@niu.edu}

\affil[1]{\orgdiv{Department of Mechanical Engineering}, \orgname{Northern Illinois University}, \orgaddress{\street{1425 W Lincoln Hwy}, \city{DeKalb}, \postcode{60115}, \state{IL}, \country{USA}}}


\abstract{Physics-informed neural networks (PINNs) provide a promising framework for solving partial differential equations while embedding the underlying physical laws directly into the learning process. This study presents a PINN-based framework for modeling transient elastodynamic wave propagation in bimaterial systems governed by the axisymmetric equations of linear elasticity. A steel-aluminum specimen representative of a Split Hopkinson Pressure Bar configuration is considered, and the governing elastodynamic equations, together with the corresponding initial, boundary, and interface conditions, are incorporated directly into the network through a physics-informed loss function. High-fidelity finite-element simulations performed using ANSYS Workbench Explicit Dynamics are used for validation and as supplementary data constraints during training. The proposed framework accurately predicts wave transmission and reflection across the bimaterial interface and reproduces axial and radial displacement histories, face-averaged responses, and the dominant stress and strain evolution with close agreement to the finite-element solutions. The trained network further demonstrates the ability to predict wave responses at previously unseen time instants and for modified material properties without requiring additional finite-element simulations, providing a continuous surrogate model for elastodynamic analysis. Mesh-sensitivity studies confirm numerical robustness, while additional material combinations demonstrate the generality of the proposed methodology. The results show that integrating physics-informed neural networks with explicit finite-element analysis provides an accurate and computationally efficient framework for elastodynamic wave propagation in heterogeneous solids, offering an effective surrogate modeling approach for high-rate solid mechanics and impact engineering applications.}

\keywords{Physics-Informed Neural Networks, Elastodynamic Wave Propagation, Bimaterial Systems, Split Hopkinson Pressure Bar, Finite Element Method, Surrogate Modeling}



\maketitle

\section{Introduction}\label{sec1}
Recent advances in scientific machine learning~\cite{Raissi_2019a, Raissi_2020a, Karniadakis_2021a, Yang_2021a, Lu_2021a, Schoenholz_2021a, Pickering_2022a, Bezgin_2023a, Zhang_2023a, Zhang_2023b, Xue_2023a, Cao_2024a, Zhang_2025a, Zhang_2026a} have created new opportunities for solving complex partial differential equations arising in computational mechanics and mechanical engineering. Among these developments, Physics-Informed Neural Networks (PINNs)~\cite{Raissi_2019a, Yang_2021a} have emerged as a promising computational framework by embedding governing equations, initial and boundary conditions directly into the neural-network training process. Unlike purely data-driven approaches, PINNs incorporate the underlying physical laws into the optimization procedure, enabling the solution of both forward problems governed by known equations and inverse problems in which unknown material properties or system parameters are inferred from limited experimental or numerical observations.

Transient wave propagation in elastic solids plays a fundamental role in structural dynamics, impact engineering, and materials science. Accurate prediction of stress-wave transmission and reflection is essential for understanding the dynamic behavior of multilayer structures, protective systems, aerospace components, and other engineering systems subjected to high-rate loading. The problem becomes considerably more challenging in heterogeneous materials, where discontinuities in material properties and imperfect interfaces strongly influence wave propagation, stress redistribution, and energy transmission. Although conventional finite-element methods provide highly accurate solutions for elastodynamic wave propagation, repeated simulations required for parametric studies, optimization, and inverse analysis are computationally expensive, motivating the development of efficient surrogate modeling techniques.

Extensive research has been devoted to analytical, numerical, and experimental investigations of elastic wave propagation. Barzkar and Adibi~\cite{Barzkar_2015a} proposed a unified viscoelastic framework connecting the Navier-Lam\'{e} and Navier-Stokes equations, providing a common description of elastic and viscous wave propagation, although their formulation was limited to one-dimensional problems. Ivanova et al.~\cite{Ivanova_2010a} investigated dynamic delamination in layered structures using a shear-lag analytical model and demonstrated the importance of interfacial stress concentrations while neglecting plastic deformation and time-dependent material behavior. Towfighi et al.~\cite{Towfighi_2002a} developed analytical models for wave propagation in anisotropic cylindrical plates, highlighting the influence of material anisotropy and geometric curvature on wave characteristics. Coker et al.~\cite{Coker_2005a} experimentally and numerically investigated frictional sliding under shear-impact loading, demonstrating transitions between crack-like and pulse-like sliding modes governed by interface friction and loading conditions. Ojha et al.~\cite{Ojha_2025a} employed LS-DYNA to study high-velocity impact of Weldox 700E steel, revealing the coupled effects of strain rate, plastic deformation, and temperature on fracture behavior. Experimental studies by Bertholf and Karnes~\cite{Bertholf_1975a} established the theoretical foundation of Split Hopkinson Pressure Bar (SHPB) testing and quantified the influence of friction, inertia, and specimen geometry on stress-wave measurements. More recently, Shin~\cite{Shin_2022a} developed a numerical solution of the Pochhammer-Chree equation to improve dispersion correction for SHPB experiments, thereby enhancing the accuracy of experimentally measured wave responses.

Parallel to these developments, Physics-Informed Neural Networks have rapidly emerged as an effective mesh-free methodology for solving partial differential equations in computational mechanics. Peng and Panesar~\cite{Peng_2024a} successfully applied PINNs to multilayer thermal simulations in additive manufacturing, demonstrating strong agreement with finite-element solutions while overcoming challenges associated with evolving geometries and material discontinuities. Wang and Thai,~\cite{Wang_2025a} integrated Classical Laminated Plate Theory with an energy-based PINN formulation to accurately predict the bending behavior of composite plates using limited training data. Margenberg et al.~\cite{Margenberg_2024a} proposed a hybrid deep neural network-multigrid framework that combines neural networks with finite-element solvers to achieve finite-element accuracy at substantially reduced computational cost. Despite these advances, relatively few studies have investigated PINN-based modeling of transient elastodynamic wave propagation in heterogeneous solids with realistic interface conditions. In particular, physics-informed formulations capable of accurately capturing stress-wave transmission and reflection across bimaterial interfaces while maintaining agreement with high-fidelity finite-element simulations remain limited.

To address this gap, the present study develops a physics-informed computational framework for transient elastodynamic wave propagation in bimaterial systems. The governing equations of linear elastodynamics are formulated for a steel-aluminum specimen with partial interfacial slip to account for frictional effects during wave transmission and reflection. The coupled partial differential equations, together with the corresponding initial, boundary, and interface conditions, are directly incorporated into a PINN framework to predict transient displacement and stress fields throughout the specimen. The proposed methodology is validated against high-fidelity finite-element simulations performed using ANSYS Workbench Explicit Dynamics under loading conditions representative of Split Hopkinson Pressure Bar experiments. The trained PINN accurately predicts displacement, stress, and strain responses, including wave propagation at previously unseen time instants, while substantially reducing the computational cost associated with repeated finite-element simulations. The proposed framework therefore provides an accurate and computationally efficient surrogate modeling approach for transient elastodynamic analysis in heterogeneous solids, with potential applications in impact-resistant structures, multilayer materials, aerospace engineering, and other high-rate solid mechanics problems.

\section{Methodology}\label{sec2}
\subsection{Governing Equations}

Transient wave propagation in the bimaterial specimen is governed by the Navier--Lam\'{e} equations for a linear, isotropic elastic solid,

\begin{equation}
\rho \frac{\partial^{2}\mathbf{u}}{\partial t^{2}}
=
(\lambda+\mu)\nabla(\nabla\cdot\mathbf{u})
+
\mu\nabla^{2}\mathbf{u},
\label{eq:navier_lame}
\end{equation}

where $\mathbf{u}$ is the displacement vector, $\rho$ is the material density, and $\lambda$ and $\mu$ are the Lam\'{e} parameters. To represent the cylindrical geometry of the Split Hopkinson Pressure Bar (SHPB) specimen, the governing equations are formulated in an axisymmetric coordinate system using the radial and axial displacement components, $u_r(r,x,t)$ and $u_x(r,x,t)$. This displacement-based formulation is employed consistently throughout the analytical model, the finite-element simulations, and the Physics-Informed Neural Network (PINN).

Under the assumption of infinitesimal deformation, the non-zero strain components in cylindrical coordinates $(r,\theta,x)$ are

\begin{equation}
\varepsilon_{rr}
=
\frac{\partial u_r}{\partial r},
\qquad
\varepsilon_{\theta\theta}
=
\frac{u_r}{r},
\qquad
\varepsilon_{xx}
=
\frac{\partial u_x}{\partial x},
\qquad
\varepsilon_{rx}
=
\frac{1}{2}
\left(
\frac{\partial u_r}{\partial x}
+
\frac{\partial u_x}{\partial r}
\right).
\label{eq:strain}
\end{equation}

The volumetric strain is

\begin{equation}
\operatorname{tr}(\boldsymbol{\varepsilon})
=
\varepsilon_{rr}
+
\varepsilon_{\theta\theta}
+
\varepsilon_{xx},
\label{eq:volstrain}
\end{equation}

where $\boldsymbol{\varepsilon}$ denotes the infinitesimal strain tensor.

For an isotropic linear elastic material, the Lam\'{e} parameters are

\begin{equation}
\lambda
=
\frac{E\nu}
{(1+\nu)(1-2\nu)},
\qquad
\mu
=
\frac{E}{2(1+\nu)},
\label{eq:lame}
\end{equation}

where $E$ is Young's modulus and $\nu$ is Poisson's ratio.

The corresponding Cauchy stress components are obtained from Hooke's law,

\begin{equation}
\begin{aligned}
\sigma_{rr}
&=
\lambda\,\operatorname{tr}(\boldsymbol{\varepsilon})
+
2\mu\varepsilon_{rr},
\\
\sigma_{\theta\theta}
&=
\lambda\,\operatorname{tr}(\boldsymbol{\varepsilon})
+
2\mu\varepsilon_{\theta\theta},
\\
\sigma_{xx}
&=
\lambda\,\operatorname{tr}(\boldsymbol{\varepsilon})
+
2\mu\varepsilon_{xx},
\\
\sigma_{rx}
&=
2\mu\varepsilon_{rx}.
\end{aligned}
\label{eq:stress}
\end{equation}

Substituting Eq.~\eqref{eq:stress} into the linear momentum equations yields the governing axisymmetric elastodynamic equations for radial and axial wave propagation~\cite{Mitchell_2017a},

\begin{equation}
\rho(x)
\frac{\partial^{2}u_r}{\partial t^{2}}
=
\frac{\partial\sigma_{rr}}{\partial r}
+
\frac{\partial\sigma_{rx}}{\partial x}
+
\frac{\sigma_{rr}-\sigma_{\theta\theta}}{r},
\label{eq:radial}
\end{equation}

\begin{equation}
\rho(x)
\frac{\partial^{2}u_x}{\partial t^{2}}
=
\frac{\partial\sigma_{rx}}{\partial r}
+
\frac{\partial\sigma_{xx}}{\partial x}
+
\frac{\sigma_{rx}}{r}.
\label{eq:axial}
\end{equation}

Within the PINN framework, the second-order temporal derivatives,

\begin{equation}
u_{r,tt}
=
\frac{\partial^{2}u_r}{\partial t^{2}},
\qquad
u_{x,tt}
=
\frac{\partial^{2}u_x}{\partial t^{2}},
\label{eq:secondtime}
\end{equation}

are evaluated using automatic differentiation. The governing equations are enforced by minimizing the physics residuals,

\begin{equation}
\mathrm{Res}_r
=
\rho u_{r,tt}
-
\left(
\frac{\partial\sigma_{rr}}{\partial r}
+
\frac{\partial\sigma_{rx}}{\partial x}
+
\frac{\sigma_{rr}-\sigma_{\theta\theta}}{r}
\right),
\label{eq:resr}
\end{equation}

\begin{equation}
\mathrm{Res}_x
=
\rho u_{x,tt}
-
\left(
\frac{\partial\sigma_{rx}}{\partial r}
+
\frac{\partial\sigma_{xx}}{\partial x}
+
\frac{\sigma_{rx}}{r}
\right),
\label{eq:resx}
\end{equation}

which are incorporated into the PINN loss function to ensure that the predicted displacement field satisfies the governing elastodynamic equations throughout the computational domain.

\subsection{Initial, Boundary, and Interface Conditions}

The governing equations are supplemented with appropriate initial, boundary, and interface conditions to ensure a well-posed elastodynamic problem.

Initially, the specimen is assumed to be undeformed and at rest,

\begin{equation}
\mathbf{u}(\mathbf{x},0)=\mathbf{0},
\qquad
\frac{\partial\mathbf{u}}{\partial t}(\mathbf{x},0)=\mathbf{0},
\label{eq:initial}
\end{equation}

where $\mathbf{u}$ denotes the displacement vector.

Axisymmetry is imposed along the cylindrical centerline $(r=0)$, while the outer cylindrical surface is assumed to be traction-free. At the incident-bar end, a prescribed axial impact velocity is applied to generate the stress wave, whereas the far end of the transmitted bar is treated as a traction-free boundary.

At material interfaces, continuity of the axial displacement is enforced,

\begin{equation}
u_x^{-}=u_x^{+},
\label{eq:dispcontinuity}
\end{equation}

where the superscripts $(-)$ and $(+)$ denote quantities evaluated on the two sides of the interface.

The interface is assumed to be frictionless; consequently, the shear traction vanishes,

\begin{equation}
\sigma_{rx}=0.
\label{eq:shearbc}
\end{equation}

To ensure physically consistent wave transmission and reflection, continuity of the axial normal stress is imposed in a weak sense across both the steel--steel and steel--aluminum interfaces,

\begin{equation}
\sigma_{xx}^{-}
=
\sigma_{xx}^{+}.
\label{eq:stresscontinuity}
\end{equation}

These initial, boundary, and interface conditions are incorporated into the PINN loss function through corresponding penalty terms, ensuring that the predicted displacement field satisfies both the governing elastodynamic equations and the prescribed physical constraints throughout the computational domain.

\subsection{Finite-Element Model}

\begin{figure}[h]
\centering
\includegraphics[width=0.9\textwidth]{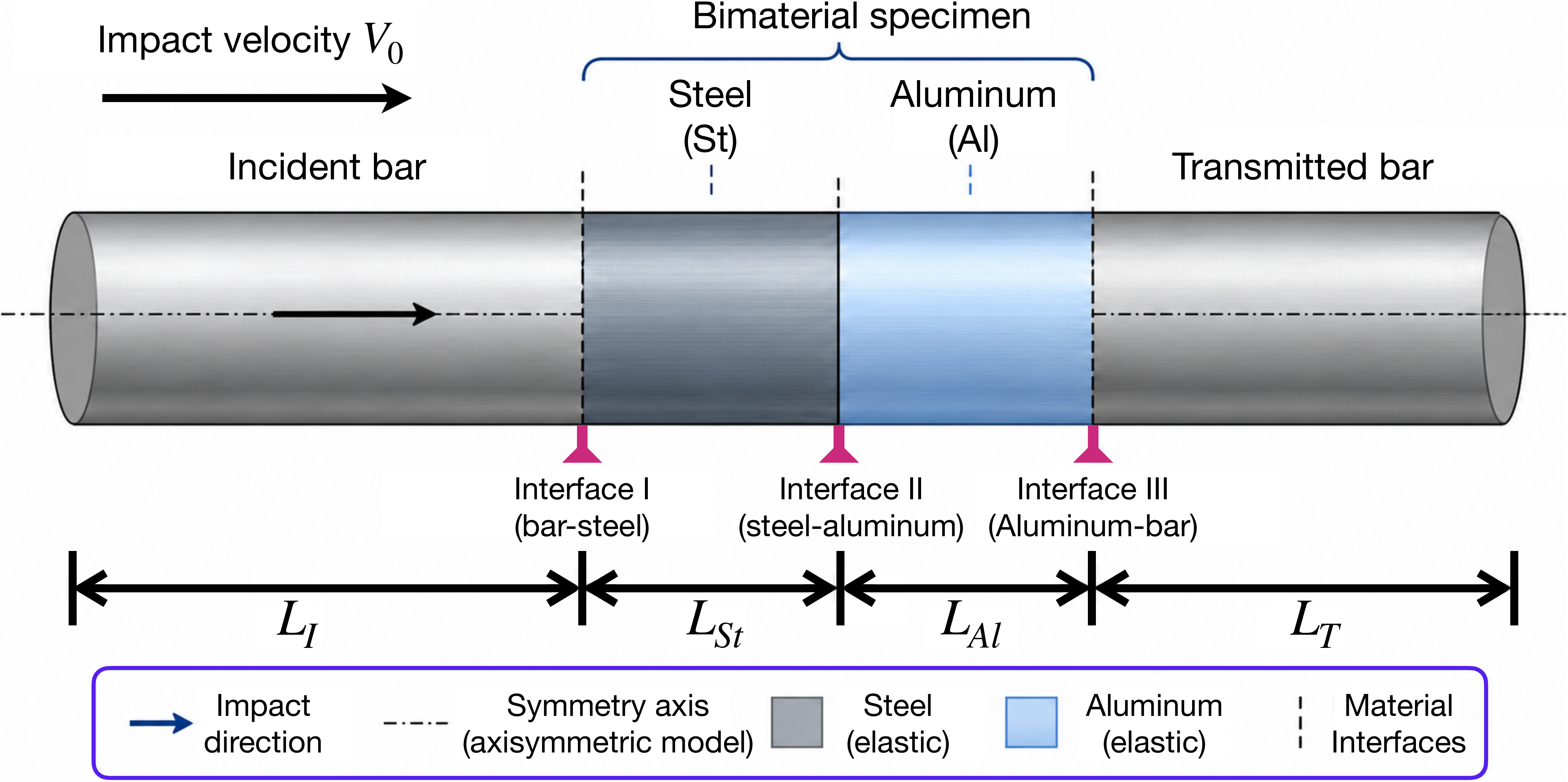}
\caption{Schematic of the Split Hopkinson Pressure Bar (SHPB) assembly consisting of an incident bar, a bimaterial steel--aluminum specimen, and a transmitted bar. The incident impact generates a compressive stress wave that propagates through the specimen and across the material interfaces.}\label{fig:schematic}
\end{figure}

\begin{figure}[h]
\centering
\includegraphics[width=1.\textwidth]{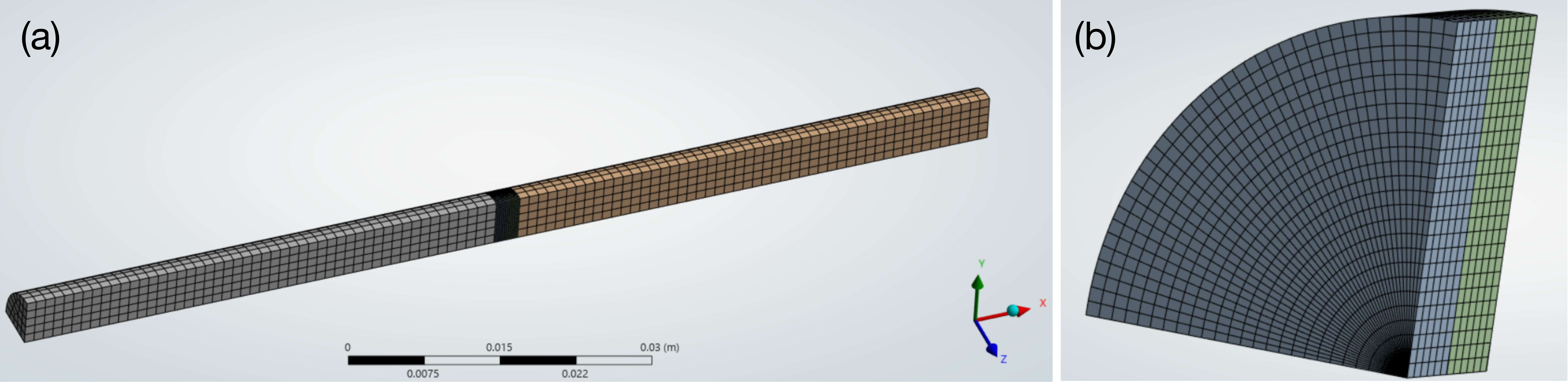}
\caption{Finite-element model of the Split Hopkinson Pressure Bar (SHPB) assembly. (a) Three-dimensional quarter-symmetry model consisting of the incident bar, bimaterial steel--aluminum specimen, and transmitted bar. (b) Refined finite-element mesh in the bimaterial specimen, showing the discretization adopted near the steel--aluminum interface to accurately resolve stress-wave transmission and reflection.}\label{fig:bar}
\end{figure}

The finite-element model is developed in ANSYS Workbench Explicit Dynamics to provide a high-fidelity reference solution for validating the proposed Physics-Informed Neural Network (PINN). All components of the Split Hopkinson Pressure Bar (SHPB) assembly are modeled as linear, isotropic elastic materials, consistent with the assumptions of the governing Navier--Lam\'{e} equations. The incident and transmitted bars are constructed from steel with a Young's modulus of $2.0\times10^{5}$~MPa, a Poisson's ratio of 0.30, and a density of $7850~\mathrm{kg\,m^{-3}}$. The bimaterial specimen consists of steel and aluminum, where the aluminum is assigned a Young's modulus of $71\,000$~MPa, a Poisson's ratio of 0.33, and a density of $2780~\mathrm{kg\,m^{-3}}$. These material properties define the spatially varying fields $\rho(x)$, $\lambda(x)$, and $\mu(x)$ in the governing elastodynamic equations, enabling material discontinuities across the steel--aluminum interface while preserving displacement continuity.

The numerical model consists of an incident bar, a bimaterial specimen composed of steel and aluminum, and a transmitted bar, as illustrated schematically in Fig.~\ref{fig:schematic}. The incident and transmitted bars each have a length of 50~mm and a diameter of 10~mm, while the bimaterial specimen has a total length of 2.5~mm with the same diameter, ensuring geometric compatibility throughout the SHPB assembly. Owing to geometric symmetry, a three-dimensional quarter-symmetry model is employed (Fig.~\ref{fig:bar}(a)), which substantially reduces the computational cost while remaining fully consistent with the axisymmetric formulation presented in Section~2.1.

Spatial discretization is designed to accurately resolve transient stress-wave propagation while maintaining computational efficiency. The incident and transmitted bars are discretized using a uniform element size of 1.0~mm, which provides sufficient resolution for one-dimensional axial wave propagation~\cite{Achenbach_2012a}. To accurately capture the steep stress and displacement gradients near the steel--aluminum interface, the bimaterial specimen is discretized using a refined element size of 0.25~mm. Figure~\ref{fig:bar}(a) shows the three-dimensional quarter-symmetry finite-element model, while Fig.~\ref{fig:bar}(b) presents the refined mesh adopted in the specimen region. The locally refined discretization enables accurate resolution of stress-wave transmission and reflection at the material interface while maintaining numerical stability during explicit time integration.

\subsection{Physics-Informed Neural Network}

\begin{figure}[h]
\centering
\includegraphics[width=1.\textwidth]{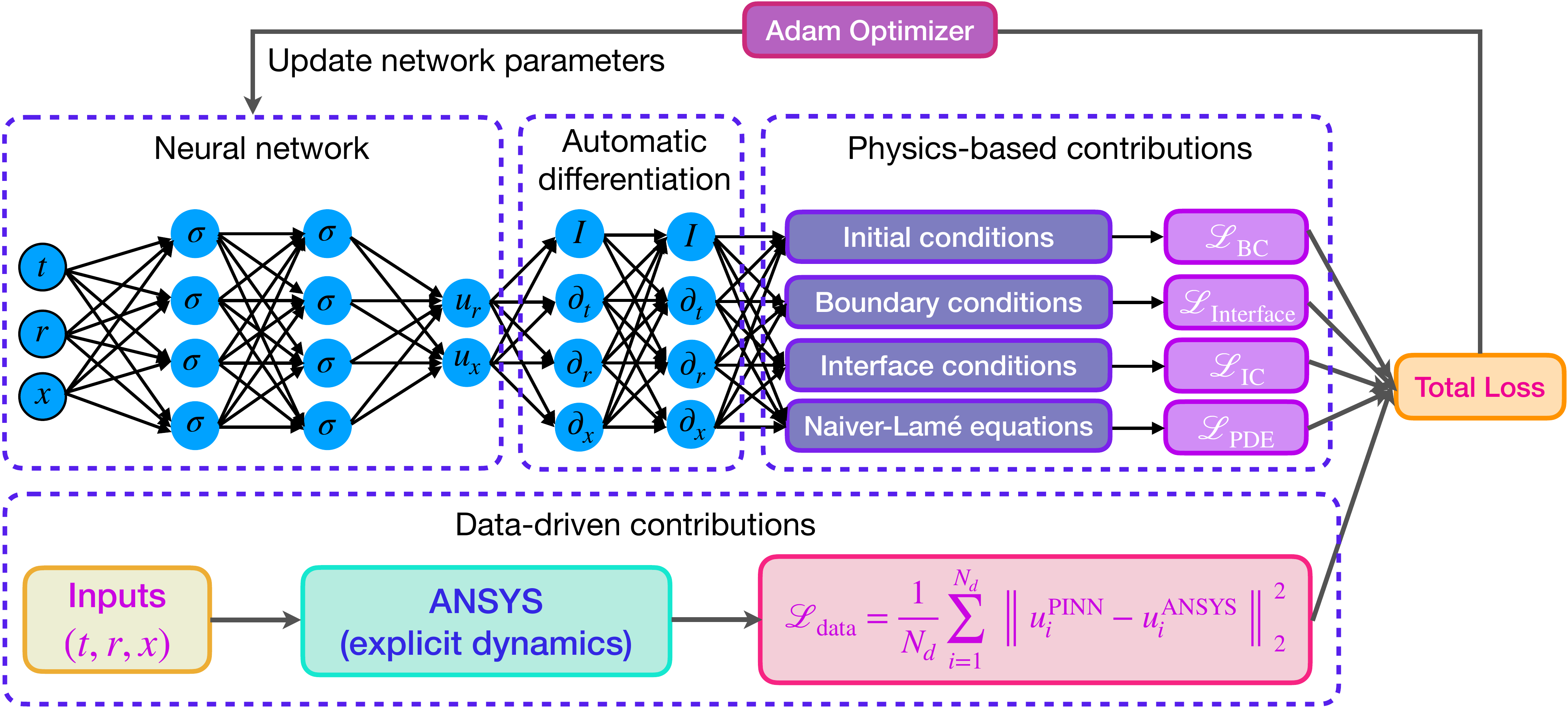}
\caption{Schematic of the proposed Physics-Informed Neural Network (PINN) framework for axisymmetric elastodynamic wave propagation. The neural network maps the spatial--temporal coordinates $(r,x,t)$ to the radial and axial displacement fields $(u_r,u_x)$. Automatic differentiation is used to enforce the governing Navier--Lam\'{e} equations together with the initial, boundary, and interface conditions through physics-based loss terms, while displacement data obtained from ANSYS Explicit Dynamics provide additional data-driven constraints. The total weighted loss is minimized using the Adam optimizer to update the network parameters.}\label{fig:pinn}
\end{figure}

The coupled axisymmetric elastodynamic equations, together with the prescribed initial, boundary, and interface conditions, constitute a high-dimensional constrained boundary-value problem for transient wave propagation in heterogeneous solids. Owing to the heterogeneous material properties and interfacial constraints, obtaining analytical solutions is generally intractable~\cite{Raissi_2019a}. To address this challenge, a Physics-Informed Neural Network (PINN) is employed as a physics-constrained surrogate model. Unlike purely data-driven neural networks, the PINN incorporates the governing physical laws directly into the training process by embedding the elastodynamic equations and all associated constraints within a unified loss function. In addition, displacement data obtained from ANSYS Explicit Dynamics are incorporated as soft constraints to improve predictive accuracy while maintaining physical consistency. Consequently, the trained PINN provides a continuous approximation of the displacement field throughout the space--time domain and can be evaluated at arbitrary spatial locations and time instants without requiring additional finite-element simulations.

\subsubsection{Network Architecture}
The PINN is implemented as a fully connected feed-forward neural network that learns the nonlinear mapping

\begin{equation}
(r,x,t)
\longrightarrow
\left(u_r(r,x,t),\,u_x(r,x,t)\right),
\label{eq:pinn_mapping}
\end{equation}

where $r$, $x$, and $t$ denote the radial coordinate, axial coordinate, and time, respectively, and $u_r(r,x,t)$ and $u_x(r,x,t)$ represent the corresponding radial and axial displacement components. The computational domain is defined by

\[
0\le r\le R,\qquad
0\le x\le L_x,\qquad
0\le t\le T,
\]

where $R$ is the radius of the cylindrical specimen, $L_x$ is its axial length, and $T$ is the total simulation time.

The network takes the spatial and temporal coordinates $(r,x,t)$ as inputs and predicts the corresponding radial and axial displacement components, which directly correspond to the nodal displacement fields obtained from ANSYS Explicit Dynamics. All required spatial and temporal derivatives are computed through automatic differentiation, allowing direct evaluation of strains, stresses, and governing-equation residuals without numerical differentiation.

Figure~\ref{fig:pinn} illustrates the overall workflow of the proposed Physics-Informed Neural Network (PINN) framework. The spatial--temporal coordinates $(r,x,t)$ are provided as inputs to a fully connected feed-forward neural network, which predicts the radial and axial displacement fields $(u_r,u_x)$. Automatic differentiation is then employed to compute the spatial and temporal derivatives required for evaluating the governing Navier--Lam\'{e} equations and constructing the residuals of the elastodynamic partial differential equations. The physics-based loss consists of contributions from the governing equations together with the prescribed initial, boundary, and interface conditions. In parallel, displacement data obtained from ANSYS Explicit Dynamics are incorporated as data-driven constraints through a data-loss term. The physics-based and data-driven contributions are combined to form the total weighted loss function, which is minimized using the Adam optimizer to iteratively update the network parameters until convergence.

\subsubsection{Physics-Informed Loss Function}

The network parameters are obtained by minimizing a composite loss function consisting of physics-based and data-driven contributions,

\begin{equation}
\mathcal{L}
=
w_f\mathcal{L}_{\mathrm{PDE}}
+
w_b\mathcal{L}_{\mathrm{BC}}
+
w_i\mathcal{L}_{\mathrm{IC}}
+
w_{\Gamma}\mathcal{L}_{\mathrm{Interface}}
+
w_d\mathcal{L}_{\mathrm{Data}},
\label{eq:loss}
\end{equation}

where $\mathcal{L}$ denotes the total loss function; $\mathcal{L}_{\mathrm{PDE}}$, $\mathcal{L}_{\mathrm{BC}}$, $\mathcal{L}_{\mathrm{IC}}$, $\mathcal{L}_{\mathrm{Interface}}$, and $\mathcal{L}_{\mathrm{Data}}$ represent the loss contributions associated with the governing partial differential equations, boundary conditions, initial conditions, interface conditions, and finite-element displacement data, respectively. The weighting coefficients $w_f$, $w_b$, $w_i$, $w_{\Gamma}$, and $w_d$ control the relative importance of the corresponding loss terms during network training. All loss terms are evaluated at collocation points distributed throughout the computational space--time domain.

The physics loss is constructed from the mean-squared residuals of the governing axisymmetric elastodynamic equations evaluated at interior collocation points $(r_f,x_f,t_f)$. The stress components $\sigma_{rr}$, $\sigma_{\theta\theta}$, $\sigma_{xx}$, and $\sigma_{rx}$ are computed from the network-predicted displacement field using the constitutive relations presented in Section~2.1 together with the piecewise-constant material properties corresponding to steel and aluminum. All spatial and temporal derivatives required for evaluating the governing equations are computed through automatic differentiation~\cite{Haghighat_2021a}. The PDE loss is defined as the mean-squared value of the normalized residuals over all interior collocation points.

Additional penalty terms enforce the prescribed initial, boundary, and interface conditions. The initial conditions constrain the displacement and velocity fields at $t=0$, while the boundary-condition loss enforces axisymmetry, traction-free boundaries, and the prescribed impact loading. Interface conditions impose continuity of the axial displacement together with the corresponding traction conditions across the steel--steel and steel--aluminum interfaces.

To further improve predictive accuracy, displacement histories obtained from ANSYS Explicit Dynamics are incorporated into the loss function as soft constraints. These observations include radial displacement histories on the steel and aluminum specimen faces together with axial displacement histories along the centerline $(r=0)$ at multiple spatial locations and time instants. For each data set, the corresponding spatial coordinates and time instants are supplied to the network, and the discrepancy between the predicted and finite-element displacement fields is minimized using a normalized mean-squared error. These data constraints anchor the learned solution to the high-fidelity finite-element simulations while preserving consistency with the governing physical laws.

\subsubsection{Training Strategy}

The network parameters are optimized by minimizing the total weighted loss function using the Adam optimizer. To improve convergence and numerical stability, interior collocation points are periodically resampled during training to enhance spatial coverage and reduce overfitting. A progressive weighting strategy is adopted in which the governing-equation residual initially receives a relatively small weight, allowing the network to first satisfy the displacement observations and boundary conditions before gradually enforcing the governing physics. Furthermore, all physical variables are appropriately normalized using suitable reference scales to improve numerical conditioning and training stability.

After convergence, the trained PINN provides a smooth and differentiable approximation of the displacement field,

\[
\left(u_r(r,x,t),\,u_x(r,x,t)\right),
\]

throughout the entire computational domain. The resulting surrogate model accurately reproduces the finite-element solution while enabling efficient interpolation and prediction at arbitrary spatial locations and time instants without repeated finite-element analyses.

\section{Results}\label{sec3}
\subsection{Validation Against ANSYS Explicit Dynamics}

\begin{figure}[h]
\centering
\includegraphics[width=\textwidth]{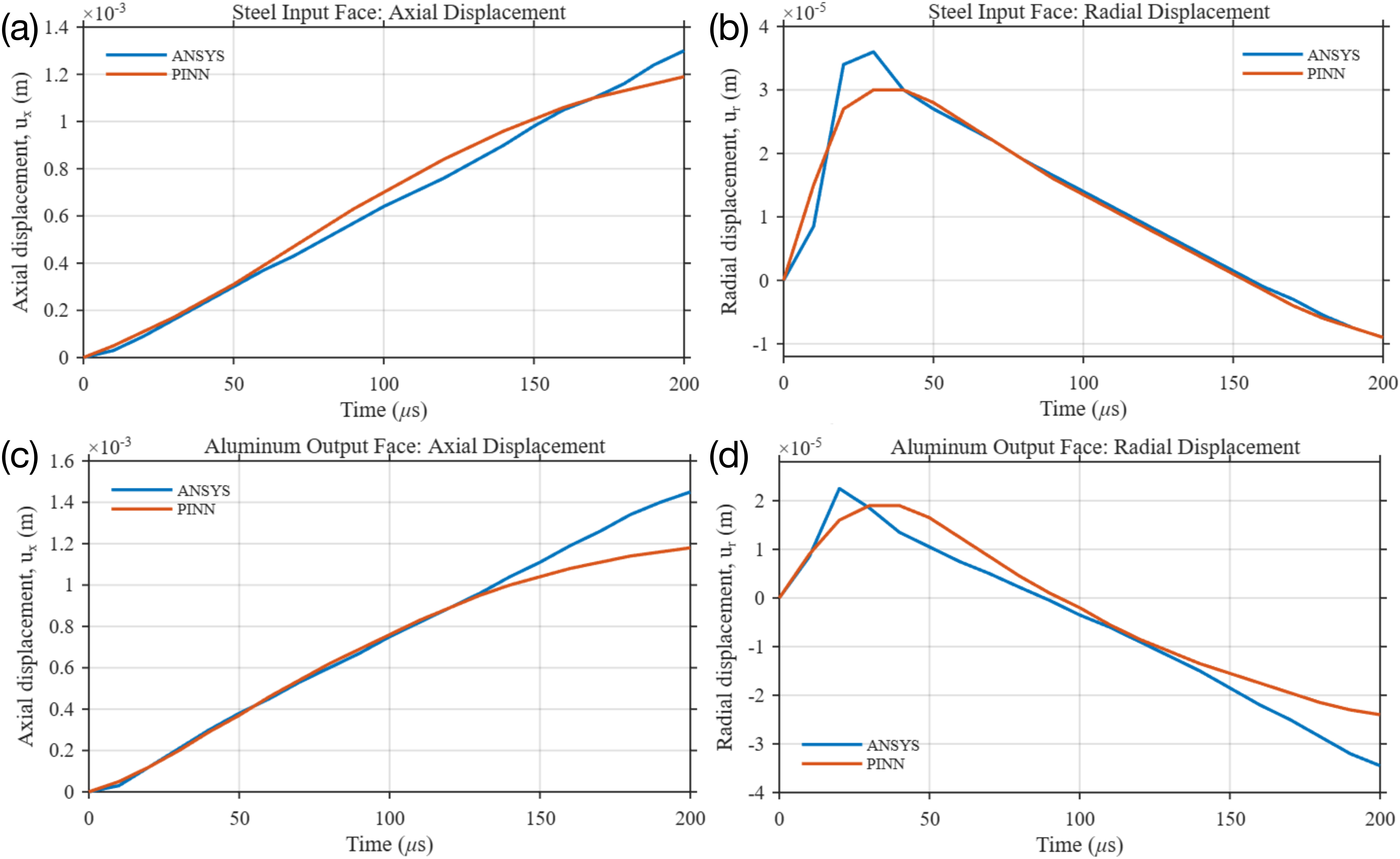}
\caption{Comparison between the displacement histories predicted by the proposed Physics-Informed Neural Network (PINN) and ANSYS Explicit Dynamics at representative nodes located on the steel input face and aluminum output face. (a) Axial displacement at the steel input-face node. (b) Radial displacement at the steel input-face node. (c) Axial displacement at the aluminum output-face node. (d) Radial displacement at the aluminum output-face node.}
\label{fig:center_node}
\end{figure}

The predictive capability of the proposed Physics-Informed Neural Network (PINN) was evaluated by comparison with high-fidelity finite-element simulations performed using ANSYS Explicit Dynamics. Representative monitoring nodes located on the steel input face and aluminum output face were selected to assess the transient elastodynamic response of the bimaterial specimen. The corresponding axial and radial displacement histories obtained from ANSYS Explicit Dynamics were incorporated into the data-loss term during training, enabling the network to simultaneously satisfy the governing elastodynamic equations while remaining consistent with the finite-element solution.

Figure~\ref{fig:center_node} compares the displacement histories predicted by the PINN with the ANSYS Explicit Dynamics results at the representative input- and output-face nodes. The axial displacement responses, shown in Figs.~\ref{fig:center_node}(a) and \ref{fig:center_node}(c), exhibit excellent agreement throughout the transient loading process. The PINN accurately captures the wave arrival time, peak displacement, and subsequent elastic unloading in both the steel and aluminum regions. Minor deviations are observed at later times, particularly at the aluminum output face, where multiple wave reflections and material impedance mismatch produce increasingly complex wave interactions.

The corresponding radial displacement histories, presented in Figs.~\ref{fig:center_node}(b) and \ref{fig:center_node}(d), are also accurately reproduced by the proposed PINN. The network successfully captures the transient radial deformation associated with Poisson coupling, including the expansion, contraction, and overall temporal evolution of the displacement field. Slight discrepancies appear after the peak response, especially at the aluminum output face, owing to its lower elastic modulus and higher compliance, which increase the sensitivity of the radial response to reflected waves and interface interactions.

Overall, the proposed PINN demonstrates excellent agreement with the ANSYS Explicit Dynamics solutions for both axial and radial displacement histories at the representative monitoring locations. The results confirm that the network accurately captures the essential characteristics of stress-wave propagation, transmission, and reflection in the steel--aluminum bimaterial specimen, demonstrating its effectiveness as a reliable physics-constrained surrogate model for axisymmetric elastodynamic wave propagation.

\subsection{Generalization to Unseen Time Instants}

\begin{figure}[h]
\centering
\includegraphics[width=\textwidth]{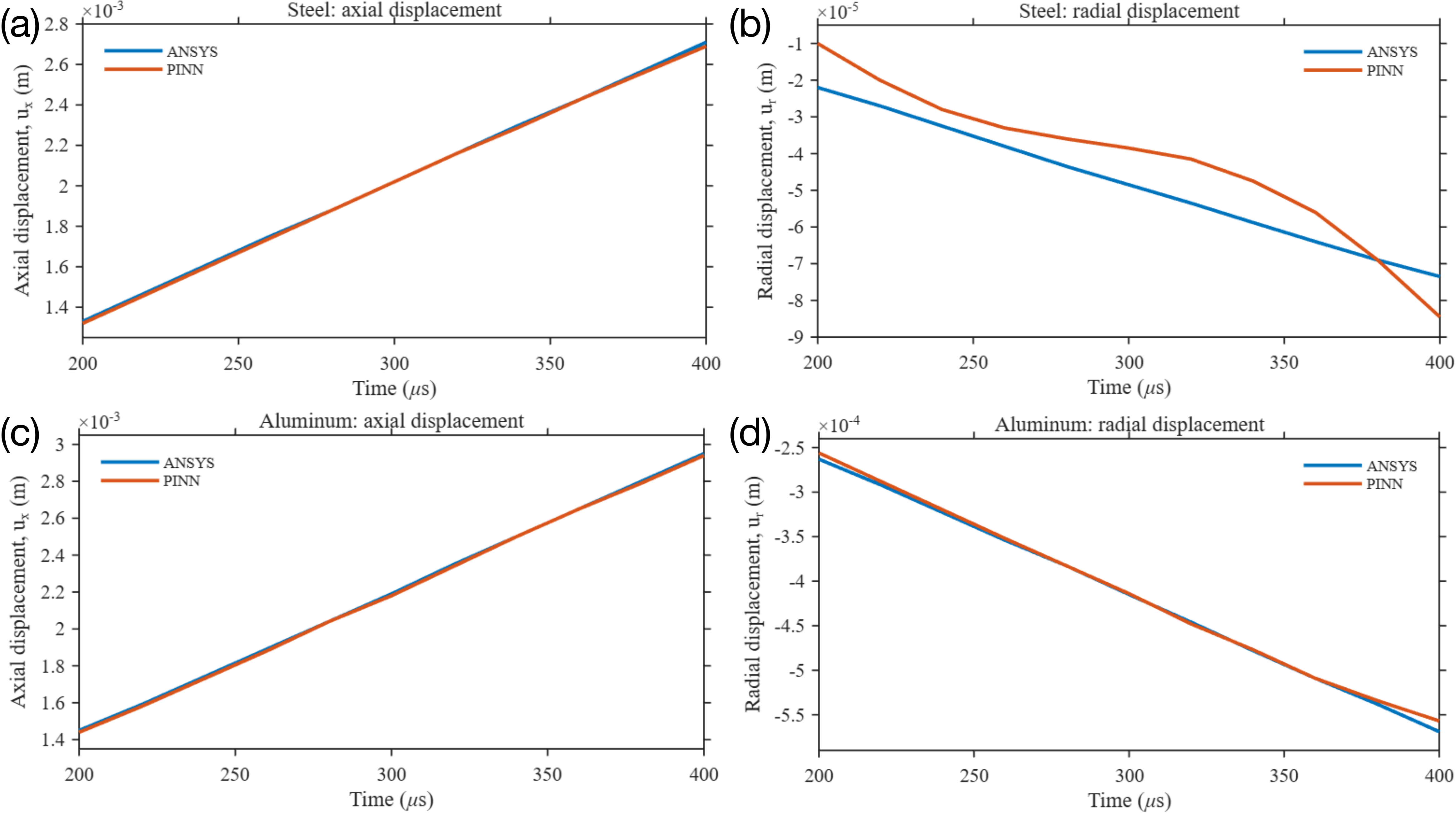}
\caption{Comparison between the displacement histories predicted by the proposed Physics-Informed Neural Network (PINN) and ANSYS Explicit Dynamics over the unseen time interval of 200--400~$\mu$s. (a) Axial displacement of the representative steel node. (b) Radial displacement of the representative steel node. (c) Axial displacement of the representative aluminum node. (d) Radial displacement of the representative aluminum node. The PINN accurately reproduces the ANSYS predictions for both materials, demonstrating excellent temporal generalization beyond the training interval.}
\label{fig:unseen}
\end{figure}

To evaluate the temporal generalization capability of the proposed Physics-Informed Neural Network (PINN), the trained model was tested over an unseen time interval of 200--400~$\mu$s without additional training. Figure~\ref{fig:unseen} compares the PINN predictions with the corresponding ANSYS Explicit Dynamics results for representative monitoring nodes in the steel and aluminum specimens.

As shown in Fig.~\ref{fig:unseen}(a) and (c), the PINN accurately reproduces the axial displacement histories throughout the unseen time interval. The predicted responses exhibit a smooth continuation beyond the training window, preserving the overall displacement evolution and demonstrating stable long-term prediction. Excellent agreement is observed for both the steel and aluminum nodes, indicating that the learned model successfully captures the underlying elastodynamic behavior rather than merely interpolating the training data.

The corresponding radial displacement histories are presented in Fig.~\ref{fig:unseen}(b) and (d). The PINN successfully captures the transient radial deformation associated with Poisson coupling, accurately reproducing the overall temporal evolution of the displacement field. Excellent agreement is observed for the aluminum node throughout the entire prediction interval. For the steel node, however, a more noticeable deviation develops near the end of the unseen time window, where the PINN overestimates the magnitude of the radial displacement. This discrepancy is likely attributable to the increased sensitivity of the radial response to late-time wave reflections, material impedance mismatch, and the accumulation of prediction errors during long-term extrapolation.

Overall, the proposed PINN demonstrates strong temporal generalization beyond the training interval. Despite the increased complexity of the late-time radial response in the steel specimen, the network accurately predicts the dominant displacement evolution in both materials while preserving the essential physics of transient wave propagation. Once trained, the PINN provides a continuous surrogate model that can be efficiently evaluated at arbitrary spatial locations and time instants without additional finite-element simulations, thereby significantly reducing the computational cost of repeated elastodynamic analyses.

\subsection{Axisymmetric Face-Averaged Responses}

\begin{figure}[h]
\centering
\includegraphics[width=\textwidth]{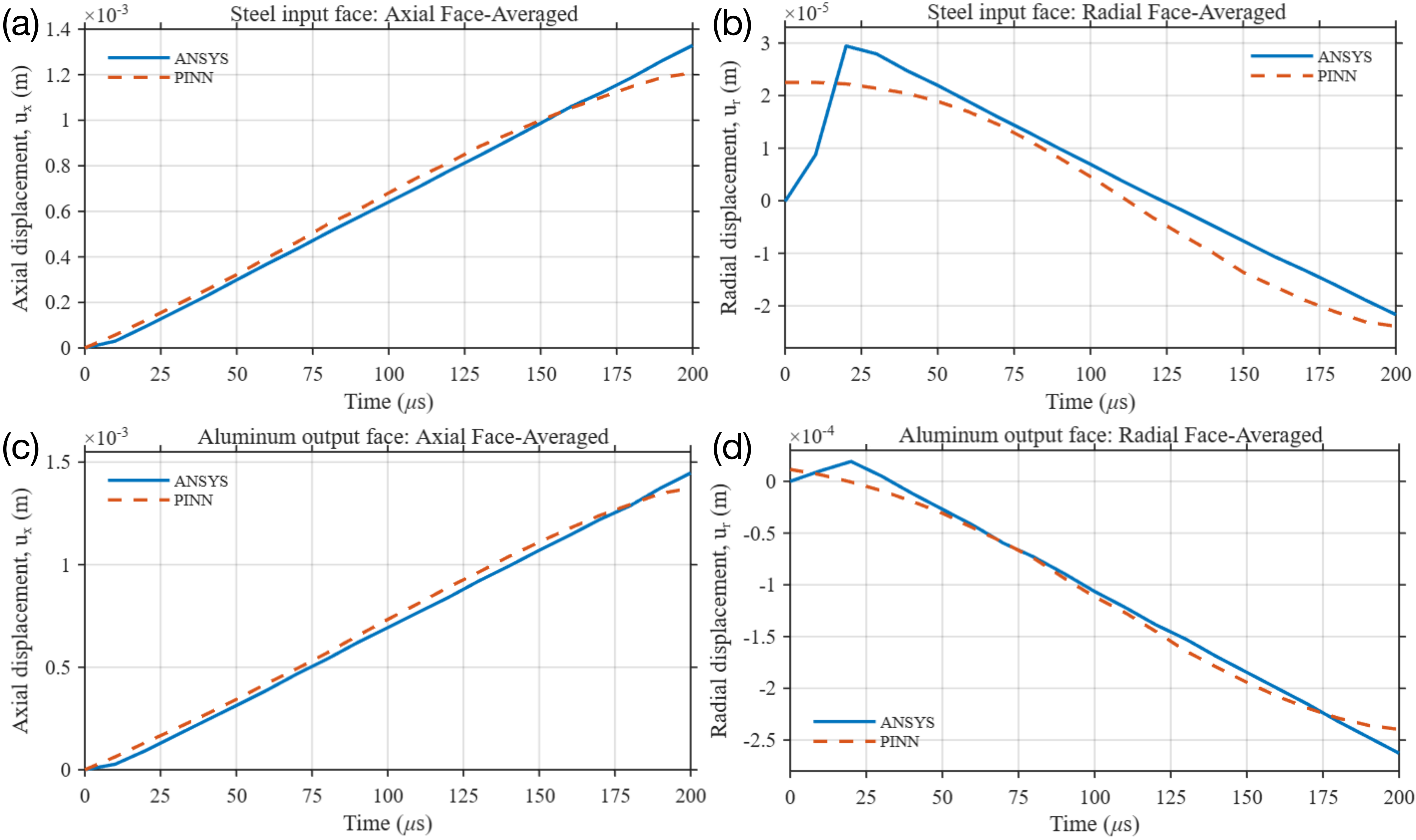}
\caption{Comparison between the face-averaged displacement histories predicted by the proposed Physics-Informed Neural Network (PINN) and ANSYS Explicit Dynamics. (a) Axial displacement at the steel input face. (b) Radial displacement at the steel input face. (c) Axial displacement at the aluminum output face. (d) Radial displacement at the aluminum output face.}
\label{fig:face_average}
\end{figure}

To further validate the proposed Physics-Informed Neural Network (PINN), face-averaged displacement histories were compared with the corresponding ANSYS Explicit Dynamics results. Face averaging is employed because Split Hopkinson Pressure Bar (SHPB) theory assumes that the measured stress and strain represent cross-sectional average quantities. Accordingly, the PINN predictions are post-processed using the same axisymmetric averaging procedure as the finite-element solution to enable a direct and physically meaningful comparison.

For an axisymmetric circular cross-section, the face-averaged displacement is computed as

\begin{equation}
\bar{u}(x,t)
=
\frac{\displaystyle\int_{0}^{R}u(r,x,t)\,r\,dr}
{\displaystyle\int_{0}^{R}r\,dr},
\label{eq:face_average}
\end{equation}

where $\bar{u}$ denotes the face-averaged displacement, $u$ represents either the axial displacement $u_x$ or the radial displacement $u_r$, and $R$ is the specimen radius. The weighting factor $r$ arises from the axisymmetric area element $dA=2\pi r\,dr$. The integral is evaluated numerically using Monte Carlo integration with 2048 sampling points distributed according to a square-root radial distribution to ensure uniform area coverage over the circular cross-section.

Figure~\ref{fig:face_average} compares the face-averaged displacement histories predicted by the PINN with the corresponding ANSYS Explicit Dynamics results. The axial displacement histories shown in Fig.~\ref{fig:face_average}(a) and (c) exhibit excellent agreement for both the steel input face and the aluminum output face. The PINN accurately captures the wave arrival, displacement evolution, and subsequent elastic response throughout the loading process.

The corresponding radial displacement histories are presented in Fig.~\ref{fig:face_average}(b) and (d). The PINN successfully reproduces the transient radial deformation associated with Poisson coupling and captures the overall temporal evolution of the cross-sectional response. Compared with the nodal predictions presented in Section~5.2, the face-averaged responses exhibit smoother behavior and improved agreement with the finite-element solutions because local numerical fluctuations are reduced through cross-sectional averaging. Minor discrepancies remain during the late stages of the steel response, where multiple wave reflections and material impedance mismatch produce increasingly complex transverse deformation.

Overall, the face-averaged results demonstrate that the proposed PINN accurately reproduces the cross-sectional elastodynamic response of the bimaterial specimen. The excellent agreement with the ANSYS Explicit Dynamics simulations confirms the capability of the proposed framework to predict the response quantities most relevant to SHPB experiments while preserving the underlying transient wave-propagation physics.

\subsection{Stress--Strain Validation}

\begin{figure}[h]
\centering
\includegraphics[width=\textwidth]{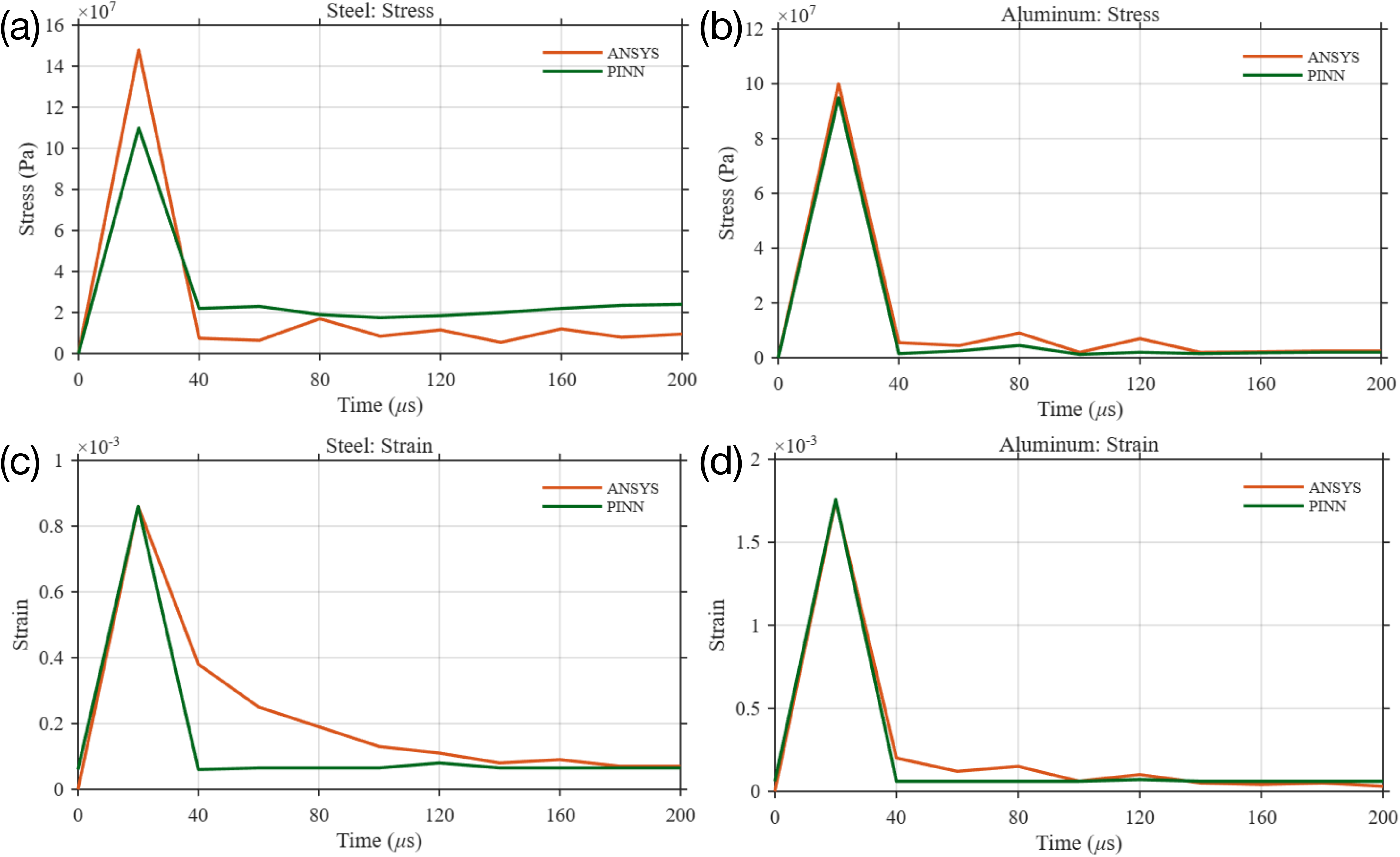}
\caption{Comparison between the stress and strain histories predicted by the proposed Physics-Informed Neural Network (PINN) and ANSYS Explicit Dynamics. (a) Stress history in the steel specimen. (b) Stress history in the aluminum specimen. (c) Strain history in the steel specimen. (d) Strain history in the aluminum specimen.}
\label{fig:stress}
\end{figure}

Following the displacement validation, the stress and strain responses predicted by the proposed Physics-Informed Neural Network (PINN) were compared with the corresponding ANSYS Explicit Dynamics results to further assess the model's ability to reproduce the transient elastodynamic behavior of the bimaterial specimen. In addition to the kinematic response, this comparison evaluates the constitutive response, stress-wave transmission, and material-interface effects.

Figure~\ref{fig:stress} compares the stress and strain histories obtained from the PINN and ANSYS Explicit Dynamics. The stress responses shown in Fig.~\ref{fig:stress}(a) and (b) indicate that both methods accurately capture the dominant compressive stress peak at approximately 20~$\mu$s, corresponding to the arrival of the first stress wave generated by the impact loading. Excellent agreement is observed for the aluminum specimen throughout the simulation. For the steel specimen, the initial peak is reproduced satisfactorily; however, larger discrepancies develop after the first wave passage, where the PINN predicts a smoother post-peak response than the finite-element solution. These differences are likely associated with the increased influence of wave reflections, stress redistribution, and interface interactions during the later stages of the transient response.

The corresponding strain histories are presented in Fig.~\ref{fig:stress}(c) and (d). Both the PINN and ANSYS accurately predict the initial transient strain peak associated with the arrival of the compressive wave. As expected, the aluminum specimen exhibits a larger strain amplitude than the steel specimen because its lower Young's modulus results in greater deformation under comparable dynamic loading. Excellent agreement is maintained for the aluminum specimen throughout the simulation. For the steel specimen, the PINN captures the initial deformation accurately but predicts a more rapid post-peak strain relaxation with reduced oscillatory behavior compared with the finite-element solution.

Overall, the proposed PINN accurately reproduces the dominant stress and strain responses during the primary loading stage while maintaining excellent agreement for the aluminum specimen over the entire simulation. Although noticeable discrepancies develop in the post-peak response of the steel specimen, the PINN successfully captures the principal constitutive behavior and transient wave propagation in the bimaterial SHPB system, demonstrating its effectiveness as a physics-constrained surrogate model for elastodynamic analysis.

\section{Conclusion}\label{sec13}
This study presented an integrated analytical, finite-element, and Physics-Informed Neural Network (PINN) framework for modeling transient wave propagation in a steel--aluminum bimaterial Split Hopkinson Pressure Bar (SHPB) specimen. The governing axisymmetric elastodynamic equations were formulated using displacement-based linear elasticity and embedded within a PINN framework, while ANSYS Explicit Dynamics provided high-fidelity reference solutions for training and validation.

The proposed PINN accurately reproduced the transient displacement response of the bimaterial system. Excellent agreement with the finite-element results was obtained for the axial displacement histories at representative monitoring locations, including the centerline, off-axis regions, and material interface. The network also successfully captured the coupled axial--radial deformation associated with Poisson effects and reproduced the cross-sectional face-averaged responses required for SHPB analysis. Furthermore, evaluation at previously unseen time instants demonstrated that the trained PINN provides smooth and physically consistent predictions beyond the discrete training data, confirming its capability as a continuous space--time surrogate model.

Stress and strain comparisons further demonstrated that the PINN accurately captured the primary compressive wave, including the wave-arrival time, peak response, and the material-dependent deformation behavior of both steel and aluminum. Although larger discrepancies were observed during the post-peak response, particularly in the steel specimen where multiple wave reflections and interface interactions become increasingly significant, the dominant elastodynamic behavior was reproduced with good accuracy. These results indicate that displacement predictions are generally more robust than derivative-based quantities such as stress and strain, highlighting the importance of validating PINN models using multiple physical fields rather than displacement alone.

Overall, the proposed framework combines the physical consistency of analytical elastodynamics, the accuracy of explicit finite-element simulations, and the computational efficiency of physics-informed neural networks. Once trained, the PINN provides rapid, mesh-free prediction of displacement, stress, and strain throughout the space--time domain, substantially reducing the need for repeated finite-element simulations in parametric studies and engineering analyses.

Future work will extend the proposed framework to other bimaterial combinations with different elastic properties and acoustic impedance contrasts in order to investigate wave transmission, reflection, and interface effects in heterogeneous materials. In addition, stronger physics enforcement, adaptive collocation strategies, and improved treatment of late-time wave reflections will be explored to further enhance prediction accuracy for complex transient elastodynamic problems.

\backmatter


\bmhead{Acknowledgements}








\bibliography{refs}

@article{Haghighat_2021a,
  title={A physics-informed deep learning framework for inversion and surrogate modeling in solid mechanics},
  author={Haghighat, Ehsan and Raissi, Maziar and Moure, Adrian and Gomez, Hector and Juanes, Ruben},
  journal={Comput.~Methods~Appl.~Mech.~Eng.},
  volume={379},
  pages={113741},
  year={2021},
  publisher={Elsevier}
}

@book{Achenbach_2012a,
  author    = {Jan D. Achenbach},
  title     = {Wave Propagation in Elastic Solids},
  publisher = {Elsevier},
  address   = {Amsterdam},
  year      = {2012},
}

@article{Mitchell_2017a,
  title={A new model for calculating the transient displacement field within a linear elastic isotropic solid with a through hole under dynamic impact: A 3D model is developed and a 2D case study is examined},
  author={Mitchell, Drew and Gau, Jenn-Terng},
  journal={Eur.~J.~Mech.~A~Solid.},
  volume={66},
  pages={269--278},
  year={2017},
  publisher={Elsevier}
}

@article{Margenberg_2024a,
  title={DNN-MG: A hybrid neural network/finite element method with applications to 3D simulations of the Navier--Stokes equations},
  author={Margenberg, Nils and Jendersie, Robert and Lessig, Christian and Richter, Thomas},
  journal={Comput.~Methods~Appl.~Mech.~Eng.},
  volume={420},
  pages={116692},
  year={2024},
  publisher={Elsevier}
}

@article{Wang_2025a,
  title={A physics-informed neural network framework for laminated composite plates under bending},
  author={Wang, Weixi and Thai, Huu-Tai},
  journal={Thin-Walled~Struct.},
  volume={210},
  pages={113014},
  year={2025},
  publisher={Elsevier}
}

@article{Peng_2024a,
  title={Multi-layer thermal simulation using physics-informed neural network},
  author={Peng, Bohan and Panesar, Ajit},
  journal={Addit.~Manuf.},
  volume={95},
  pages={104498},
  year={2024},
  publisher={Elsevier}
}

@article{Shin_2022a,
  title={Pochhammer--Chree equation solver for dispersion correction of elastic waves in a (split) Hopkinson bar},
  author={Shin, Hyunho},
  journal={Proc.~Inst.~Mech.~Eng.~Part~J.~Mech.~Eng.~Sci.},
  volume={236},
  number={1},
  pages={80--87},
  year={2022},
  publisher={SAGE Publications Sage UK: London, England}
}

@article{Bertholf_1975a,
  title={Two-dimensional analysis of the split Hopkinson pressure bar system},
  author={Bertholf, LD and Karnes, CH},
  journal={J.~Mech.~Phys.~Solids},
  volume={23},
  number={1},
  pages={1--19},
  year={1975},
  publisher={Elsevier}
}

@article{Ojha_2025a,
  title={Fracture Prediction in Weldox 700E Steel Subjected to High Velocity Impact Using LS-DYNA},
  author={Ojha, Nikesh Kumar and Saxena, Ravindra K and Vashishtha, Govind and Chauhan, Sumika},
  journal={Appl.~Sci.},
  volume={15},
  number={7},
  pages={3677},
  year={2025},
  publisher={MDPI}
}

@article{Coker_2005a,
  title={Frictional sliding modes along an interface between identical elastic plates subject to shear impact loading},
  author={Coker, D and Lykotrafitis, G and Needleman, A and Rosakis, AJ},
  journal={J.~Mech.~Phys.~Solids},
  volume={53},
  number={4},
  pages={884--922},
  year={2005},
  publisher={Elsevier}
}

@article{Towfighi_2002a,
  title={Elastic wave propagation in circumferential direction in anisotropic cylindrical curved plates},
  author={Towfighi, S and Kundu, T and Ehsani, M},
  journal={J.~Appl.~Mech.},
  volume={69},
  number={3},
  pages={283--291},
  year={2002}
}

@article{Ivanova_2010a,
  title={Interface behavior of a bimaterial plate under dynamic loading},
  author={Ivanova, Jordanka and Nikolova, Gergana and Dineva, Petia and Becker, Wilfried},
  journal={ J.~Eng.~Mech.},
  volume={136},
  number={10},
  pages={1194--1201},
  year={2010},
  publisher={American Society of Civil Engineers}
}

@article{Barzkar_2015a,
  title={On the propagation of longitudinal stress waves in solids and fluids by unifying the Navier-Lame and Navier-Stokes Equations},
  author={Barzkar, Ahmad and Adibi, Hojatollah},
  journal={Math.~Probl.~Eng.},
  volume={2015},
  number={1},
  pages={789238},
  year={2015},
  publisher={Wiley Online Library}
}

@article{Zhang_2026a,
  title={A differentiable, shock-capturing neural solver for compressible flow simulation},
  author={Zhang, Bo},
  journal={Phys.~Fluids},
  volume = {38},
  number = {4},
  pages = {046105},
  year = {2026},
  month = {04},
  publisher={AIP Publishing}
}

@article{Xue_2023a,
  title={JAX-FEM: A differentiable GPU-accelerated 3D finite element solver for automatic inverse design and mechanistic data science},
  author={Xue, Tianju and Liao, Shuheng and Gan, Zhengtao and Park, Chanwook and Xie, Xiaoyu and Liu, Wing Kam and Cao, Jian},
  journal={Comput.~Phys.~Commun.},
  volume={291},
  pages={108802},
  year={2023},
  publisher={Elsevier}
}

@article{Schoenholz_2021a,
  title={JAX, MD A framework for differentiable physics},
  author={Schoenholz, Samuel S and Cubuk, Ekin D},
  journal={J. Stat. Mech.: Theory Exp.},
  volume={2021},
  number={12},
  pages={124016},
  year={2021},
  publisher={IOP Publishing}
}

@article{Bezgin_2023a,
  title={JAX-Fluids: A fully-differentiable high-order computational fluid dynamics solver for compressible two-phase flows},
  author={Bezgin, Deniz A and Buhendwa, Aaron B and Adams, Nikolaus A},
  journal={Comput.~Phys.~Commun.},
  volume={282},
  pages={108527},
  year={2023},
  publisher={Elsevier}
}

@article{Zhang_2025a,
  title={Banach neural operator for Navier-Stokes equations},
  author={Zhang, Bo},
  journal={Phys.~Fluids},
  volume={37},
  number={8},
  pages = {086166},
  year={2025},
  publisher={AIP Publishing}
}

@article{Cao_2024a,
  title={Laplace neural operator for solving differential equations},
  author={Cao, Qianying and Goswami, Somdatta and Karniadakis, George Em},
  journal={Nat.~Mach.~Intell.},
  volume={6},
  number={6},
  pages={631--640},
  year={2024},
  publisher={Nature Publishing Group UK London}
}

@article{Zhang_2023a,
  title={Nonlinear mode decomposition via physics-assimilated convolutional autoencoder for unsteady flows over an airfoil},
  author={Zhang, Bo},
  journal={Phys.~Fluids},
  volume={35},
  number={9},
  pages = {095115},
  year={2023},
  publisher={AIP Publishing}
}

@article{Pickering_2022a,
  title={Discovering and forecasting extreme events via active learning in neural operators},
  author={Pickering, Ethan and Guth, Stephen and Karniadakis, George Em and Sapsis, Themistoklis P},
  journal={Nat.~Comput.~Sci.},
  volume={2},
  number={12},
  pages={823--833},
  year={2022},
  publisher={Nature Publishing Group US New York}
}

@article{Karniadakis_2021a,
  title={Physics-informed machine learning},
  author={Karniadakis, George Em and Kevrekidis, Ioannis G and Lu, Lu and Perdikaris, Paris and Wang, Sifan and Yang, Liu},
  journal={Nat.~Rev.~Phys.},
  volume={3},
  number={6},
  pages={422--440},
  year={2021},
  publisher={Nature Publishing Group UK London}
}

@article{Lu_2021a,
  title={Learning nonlinear operators via DeepONet based on the universal approximation theorem of operators},
  author={Lu, Lu and Jin, Pengzhan and Pang, Guofei and Zhang, Zhongqiang and Karniadakis, George Em},
  journal={Nat.~Mach.~Intell.},
  volume={3},
  number={3},
  pages={218--229},
  year={2021},
  publisher={Nature Publishing Group UK London}
}

@article{Zhang_2023b,
  title={Airfoil-based convolutional autoencoder and long short-term memory neural network for predicting coherent structures evolution around an airfoil},
  author={Zhang, Bo},
  journal={Comput.~Fluids},
  volume={258},
  pages={105883},
  year={2023}
}

@article{Raissi_2020a,
  title={Hidden fluid mechanics: Learning velocity and pressure fields from flow visualizations},
  author={Raissi, Maziar and Yazdani, Alireza and Karniadakis, George Em},
  journal={Science},
  volume={367},
  number={6481},
  pages={1026--1030},
  year={2020},
  publisher={American Association for the Advancement of Science}
}

@article{Yang_2021a,
  title={B-PINNs: Bayesian physics-informed neural networks for forward and inverse PDE problems with noisy data},
  author={Yang, Liu and Meng, Xuhui and Karniadakis, George Em},
  journal={J.~Comput.~Phys.},
  volume={425},
  pages={109913},
  year={2021},
  publisher={Elsevier}
}

@article{Raissi_2019a,
  title={Physics-informed neural networks: A deep learning framework for solving forward and inverse problems involving nonlinear partial differential equations},
  author={Raissi, Maziar and Perdikaris, Paris and Karniadakis, George E},
  journal={J.~Comput.~Phys.},
  volume={378},
  pages={686--707},
  year={2019},
  publisher={Elsevier}
}

\end{document}